\documentclass[journal]{IEEEtran}

\ifCLASSINFOpdf
\else
\fi

\usepackage{cite}
\usepackage{times}
\usepackage{algorithmic}
\usepackage{array}
\usepackage{amssymb}
\usepackage{mdwmath}
\usepackage{mdwtab}
\usepackage{eqparbox}
\usepackage{cite}
\usepackage{url}
\usepackage{multicol,multirow}
\usepackage[cmex10]{amsmath}
\usepackage{makeidx}
\usepackage{diagbox}
\usepackage{rotating,soul}
\usepackage{wrapfig}
\usepackage{booktabs}
\usepackage[usenames,dvipsnames]{color}
\usepackage{ragged2e}
\usepackage{pifont}
\usepackage{overpic}
\makeindex

\usepackage{geometry}
\RequirePackage{silence}
\DeclareGraphicsExtensions{.pdf,.jpg,.png}

\usepackage[colorlinks,bookmarksnumbered,bookmarksopen,linkcolor=black,citecolor=black,urlcolor=black]{hyperref}

\def\ie{\emph{i.e.}}
\def\eg{\emph{e.g.}}
\def\etal{{\em et al.}}

\def\ours{FAP-Net}

\newcommand{\addFig}[1]{}
\newcommand{\addFigs}[1]{}

\graphicspath{{./figs/}}

\newlength\savedwidth
\newcommand{\whline}[1]{\noalign{\global\savedwidth\arrayrulewidth \global\arrayrulewidth #1}%
                   \hline \noalign{\global\arrayrulewidth\savedwidth}}

\begin{document}

\title{Feature Aggregation and Propagation Network for Camouflaged Object Detection}

\author{Tao Zhou,~\IEEEmembership{Member,~IEEE,}
        Yi Zhou,
        Chen Gong,~\IEEEmembership{Member,~IEEE,}
        Jian Yang,~\IEEEmembership{Member,~IEEE}\\
        and~Yu Zhang,~\IEEEmembership{Senior Member,~IEEE}

\thanks{T. Zhou, C. Gong, and J. Yang are with PCA Lab, Key Lab of Intelligent Perception and Systems for High-Dimensional Information of Ministry of Education, and Jiangsu Key Lab of Image and Video Understanding for Social Security, School of Computer Science and Engineering, Nanjing University of Science and Technology, Nanjing 210094, P.R. China. 
(e-mails: taozhou.ai@gmail.com; chen.gong@njust.edu.cn; csjyang@njust.edu.cn).}
\thanks{Y. Zhou is with the School of Computer Science and Engineering, Southeast University, Nanjing 211189, China (e-mail: yizhou.szcn@gmail.com).}
\thanks{Y. Zhang is with the Department of Bioengineering, Lehigh University, Bethlehem, PA 18015, USA (e-mail: yuzi20@lehigh.edu).}
\thanks{Corresponding author: Chen Gong (chen.gong@njust.edu.cn).}
}

\markboth{IEEE TRANSACTIONS ON IMAGE PROCESSING,~Vol.~xx,No.~xx,~xxx.~xxxx}%
{Liu \MakeLowercase{\textit{et al.}}: Dynamic Feature Integration for Simultaneous Detection of Salient Object, Edge and Skeleton}

\maketitle


\begin{abstract}

Camouflaged object detection (COD) aims to detect/segment camouflaged objects embedded in the environment, which has attracted increasing attention over the past decades. Although several COD methods have been developed, they still suffer from unsatisfactory performance due to the intrinsic similarities between the foreground objects and background surroundings. In this paper, we propose a novel Feature Aggregation and Propagation Network (FAP-Net) for camouflaged object detection. Specifically, we propose a Boundary Guidance Module (BGM) to explicitly model the boundary characteristic, which can provide boundary-enhanced features to boost the COD performance. To capture the scale variations of the camouflaged objects, we propose a Multi-scale Feature Aggregation Module (MFAM) to characterize the multi-scale information from each layer and obtain the aggregated feature representations. Furthermore, we propose a Cross-level Fusion and Propagation Module (CFPM). In the CFPM, the feature fusion part can effectively integrate the features from adjacent layers to exploit the cross-level correlations, and the feature propagation part can transmit valuable context information from the encoder to the decoder network via a gate unit. Finally, we formulate a unified and end-to-end trainable framework where cross-level features can be effectively fused and propagated for capturing rich context information. Extensive experiments on three benchmark camouflaged datasets demonstrate that our \ours~outperforms other state-of-the-art COD models. Moreover, our model can be extended to the polyp segmentation task, and the comparison results further validate the effectiveness of the proposed model in segmenting polyps. The source code and results will be released at \href{https://github.com/taozh2017/FAPNet}{https://github.com/taozh2017/FAPNet}.

\end{abstract}

\begin{IEEEkeywords}
Camouflaged object detection, boundary guidance module, multi-scale feature aggregation, cross-level fusion, feature propagation.

\end{IEEEkeywords}

\IEEEpeerreviewmaketitle

\section{Introduction}

\IEEEPARstart{C}{amouflaged} Object Detection (COD) aims to identify objects with a similar texture to their surroundings. Camouflaged objects can be roughly classified into two types, \ie, natural and artificial camouflaged objects. Natural camouflaged objects hide in the background environment with their own advantages (\eg, color, shape, etc.) to adapt to the environment 
\cite{le2021camouflaged,li2018fusion,chen2022camouflaged,fan2021concealed}, while artificial camouflaged objects often occur in a real-world scenario. It has a variety of applications, such as security and surveillance (\eg, search-and-rescue work \cite{mishra2020drone}), agriculture (\eg, detecting agricultural pests), and medical imaging analysis (\eg, lung infection segmentation \cite{fan2020inf}, and polyp segmentation \cite{he2018hookworm}). Therefore, due to its application and scientific value, COD has attracted more and more attention.

Compared to generic object detection \cite{liu2020deep}, COD is a more challenging task due to the high intrinsic similarities between the camouflaged objects and their background. Camouflaged objects often have a diversity of size, color, shape, and texture, which aggravates difficulties in accurately detecting camouflaged objects. To overcome this challenge, various COD models have been developed to improve detection performance. In the early years, several traditional COD methods \cite{singh2013survey,bhajantri2006camouflage} have been proposed to segment camouflaged objects by using manually designed features. Recently, due to the development of deep learning-based representation methods, many deep learning-based COD methods have been proposed to obtain state-of-the-art performance \cite{le2019anabranch,fan2020camouflaged,wang2021d,mei2021camouflaged,ji2022fast}. For example, ANet \cite{le2019anabranch} utilizes a classification network to determine whether the image contains camouflaged objects or not, and then uses a fully convolutional network for COD. SINet \cite{fan2020camouflaged} is proposed to utilize a search module to coarsely select the candidate regions of camouflaged objects and then proposes an identification module to precisely detect camouflaged objects. More importantly, a large-scale dataset for COD also is proposed in \cite{fan2020camouflaged}, which advances this field and promotes more explorations.

Although progress has been made in the COD field, existing methods could still misunderstand camouflaged objects as the background due to their similar texture, and there is still considerable room for improving COD. \emph{First}, scale variation is one of the major challenges in the COD, how to effectively characterize the multi-scale information from a convolutional layer deserves further exploration. \emph{Second}, several COD methods often integrate multi-level features and then feed them into the decoder network, while they ignore the contributions of the feature representation from different encoder blocks. \emph{Third}, due to the boundary between a camouflaged object and its background is not sharp, thus it is helpful to locate the boundaries of camouflaged objects or incorporate boundary-attention features for improving the COD performance. 

To this end, we propose a novel COD framework, \ie, Feature Aggregation and Propagation Network (FAP-Net), to accurately detect camouflaged objects. Specifically, we first propose a Boundary Guidance Module (BGM) to learn the boundary-enhanced representations, which are then incorporated into the decoder network via a layer-wise manner to help the model detect the boundaries of camouflaged objects. Moreover, we propose a Multi-scale Feature Aggregation Module (MFAM) to exploit multi-scale information from a single convolutional block. To effectively integrate the multi-level features, we propose a Cross-level Fusion and Propagation Module (CFPM) to first fuse cross-level features. In addition, the feature propagation part can adaptively weigh the contributions of features from the encoder and decoder, which makes the decoder obtain more effective features from the encoder to boost the COD performance. Extensive experiments on three benchmark datasets demonstrate that our \ours~performs favorably against other state-of-the-art COD methods under different evaluation metrics. Moreover, the proposed model has also been extended to the polyp segmentation task, and the effectiveness can be further validated.

The main \textbf{contributions} of this paper are four-fold: 
\begin{itemize}
	\item We propose a novel \ours~for the COD task, which can effectively integrate cross-level features and propagate the valuable context information from the encoder to the decoder for accurately detecting camouflaged objects.

	\item A \emph{Boundary Guidance Module} is proposed to learn the boundary-enhanced representations, which preserve the local characteristics and boundary information of the original images to boost the COD performance. 
	
	\item We propose a \emph{Multi-scale Feature Aggregation Module} to learn the multi-scale aggregated features, which can adaptively extract multi-scale information from each level to deal with scale variations.

	\item We propose a \emph{Cross-level Fusion and Propagation Module} to effectively fuse cross-level features and propagate useful information from the encoder to the decoder network, which makes that our model can adaptively balance the contribution of the feature of each encoder block to the decoder network.

\end{itemize}

The rest of this paper is organized as follows. We discuss several related works in Section~\ref{related}. We then provide the details of the proposed \ours~in Section~\ref{Methodology}. In Section~\ref{Experiments}, we provide the experimental results and related analysis. Finally, we conclude the paper in Section~\ref{conclusion}.

\section{Related Work}
\label{related}

In this section, we present a brief overview of the three types of works that are most related to our method, including camouflaged object detection, multi-scale/level feature learning, and gated mechanism.

\subsection{Camouflaged Object Detection}

Early COD methods focused on detecting the foreground areas, and proposed several methods based on handcrafted features, including color, intensity, shape, direction, and edge \cite{singh2013survey,survey20}. For instance, Mondal \etal~\cite{mondal2017partially} proposed a tracking-by-detection strategy to discover and track camouflaged objects, in which multiple types of features (including histogram of orientation gradients, CIELab, and locally adaptive ternary pattern) are integrated to represent a camouflaged object. However, due to the limited-expression ability of handcrafted features, these early models often are unable to obtain promising performance. To address this, several deep learning based methods have been developed and obtain promising camouflaged object detection performance. For example, Li \etal~\cite{li2017} proposed a new camouflaged color target detection model based on image enhancement, in which the image enhancement algorithm is adapted to capture the difference between the target and background features. Yan \etal~\cite{yan2020} proposed to leverage both instance segmentation and adversarial attack to achieve camouflaged object segmentation, which can effectively capture different layouts of the scene to boost the segmentation performance. Lamdouar \etal~\cite{Lamdouar2020} proposed a new camouflaged object segmentation model, which consists of two components, \ie, a differentiable registration module is used to highlight object boundaries, and a motion segmentation module is used to discover moving regions. Fan \etal~\cite{fan2020camouflaged} proposed a novel and effective COD approach, termed Search Identification Network (SINet). Li \etal~\cite{li2021} proposed an enhanced cascade decoder network to identify camouflaged marine animals. Mei \etal~\cite{mei2021camouflaged} proposed a positioning and focus network to improve the COD performance.

\subsection{Multi-scale/level Feature Learning}

\textbf{Multi-scale Feature Extraction}. Multi-scale feature representations have been used for detection and segmentation tasks~\cite{fpn,guo2020augfpn,wang2019salient,zhang2019hyperfusion,pang2020multi,ding2018context}. A representative work is Feature Pyramid Network (FPN)~\cite{fpn}, which constructs multi-scale feature maps to detect objects at different scales. Guo \etal~\cite{guo2020augfpn} proposed a novel feature pyramid network to fully exploit the potential of multi-scale features. Besides, Wang \etal~\cite{wang2019salient} proposed a pyramid attention module, which obtains multi-scale attention maps to enhance feature representations using multiple downsampling and softmax operations on different positions. In \cite{zhang2019hyperfusion}, a hyper-dense fusion module is proposed to diversify the contributions of multi-scale features from local and global perspectives. Pang \etal~\cite{pang2020multi} proposed an Aggregate Interaction Module (AIM) to integrate features of adjacent levels in the encoder network. Besides, Ding~\etal~\cite{ding2018context,ding2020semantic} proposed a gated sum strategy to selectively aggregate different scale features for semantic segmentation. Different from these methods~\cite{ding2018context,ding2020semantic}, we focus on fusing cross-level features and propagating the useful information from the encoder to the decoder network, resulting in balancing the contribution of the feature of each encoder block to the decoder network. 

\begin{figure*}[t]
	\begin{centering}
		\includegraphics[width=0.95\textwidth]{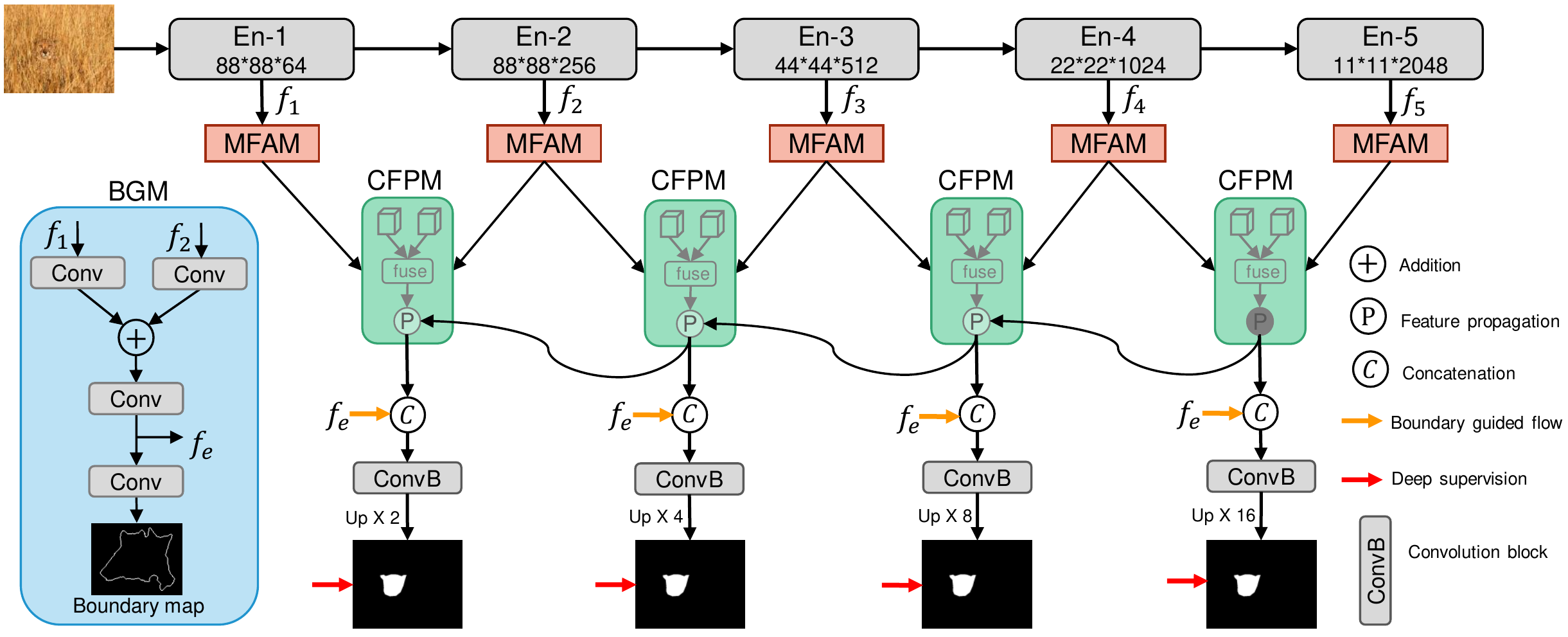}\vspace{-0.1cm}
		\caption{The overall architecture of the proposed \ours, consisting of three key components, \ie, boundary guidance module (see details in Sec.~\ref{boundary}), multi-scale feature aggregation module (see details in Sec.~\ref{aggregation}), and cross-level fusion and propagation module (see details in Sec.~\ref{fusion}).}
		\label{fig2}
	\end{centering}
\end{figure*}

\textbf{Multi-level Feature Integration}. Several works have been developed to study the integration of multi-level features. For example, in semantic segmentation \cite{fan2018multi,zhang2021prototypical,park2017rdfnet}, feature maps from selected levels are utilized with a shortcut connection to provide multiple granularities for boosting the segmentation performance. In visual recognition \cite{yang2016multilayer,cai2017higher,kuang2018integrating}, deep features from some selected layers can be merged together to improve the final layer representation. 
Besides, Zhang \etal~\cite{zhang2017amulet} proposed a generic aggregating multi-level convolutional feature framework for salient object detection, which integrates multi-level feature maps into multiple resolutions by incorporating coarse semantics and fine details. Hu \etal~\cite{hu2018recurrently} proposed to fully exploit the complementary information from multiple layers by recurrently concatenating multi-layer features to locate salient objects. 
In addition, several multi-level feature fusion strategies and multi-modal interactions have been developed and applied in several detection and
segmentation tasks~\cite{zhou2021specificity,zhang2019synthesizing,zhou2022consistency,ding2021vision}.

\subsection{Gated Mechanism}

The gated mechanism is proposed to adaptively control the flow of information and is widely applied in several computer vision tasks. For example, Cheng \etal~\cite{cheng2017locality} designed a gated fusion module to adaptively integrate the two modalities (\ie, RGB and depth) for object recognition. Zhang \etal~\cite{zhang2018bi} proposed a gated bi-directional message passing module to adaptively incorporate multi-level features. Liu \etal~\cite{liu2020cross} regarded an adaptive gated fusion module as a part of the discriminator network to adaptively integrate the RGB and depth features, which is beneficial to obtaining an effective gated fusion of saliency maps during adversarial learning. Zhao \etal~\cite{zhao2020suppress} utilized multilevel gate units to balance the contribution of each encoder path and also suppress the activation of the features from non-salient regions. Zhou \etal~\cite{zhou2020gfnet} proposed a gate fusion module to regularize the process of feature fusion, leading to obtaining better results via filtering noise and interference. Most above methods often consider the information fusion or interaction between different levels either in the encoder or decoder. We integrate the features from the encoder network and the decoder one via a gate propagation strategy, which automatically learns the contributions of different features from the encoder and decoder to boost the segmentation performance.

\section{Methodology}
\label{Methodology}

In this section, we first provide an overview of the proposed \ours~for the COD task. Then we present the three key components of our model. Finally, we present the overall loss function of the proposed COD model.

\subsection{Overview} 
\label{Overview}

Fig.~\ref{fig2} shows the overall architecture of the proposed \ours, consisting of three key components: the multi-scale feature aggregation module, cross-level fusion and propagation module, and boundary guidance module. Specifically, an image is first fed into the encoder network (Res2Net-50~\cite{pami20Res2net} as the backbone), to extract multi-level features, which are denoted as $f_i~ (i=1,2,\dots,5)$. Therefore, we have a feature resolution of $\frac{W}{4}\times\frac{H}{4}$ for the first level, and a general resolution of $\frac{W}{2^i}\times\frac{H}{2^i}$ (when $i>1$). Due to the low-level features (\ie, $f_1$ and $f_2$) containing rich boundary information, we propose a BGM to capture the boundaries and obtain the boundary-enhanced feature representation. Then, to reduce the channel size (with large computation complexity) and extract multi-scale features, the multi-level features $f_i$ are fed into the proposed MFAM to capture camouflaged objects' scale variations. After that, the aggregated features are fed into the proposed CFPM to effectively integrate cross-level features and propagate the fused features to the decoder network. More importantly, the boundary-enhanced feature representations can be also combined into the decoder network. Finally, multiple side-out supervised strategies are implemented to boost the COD performance. We will provide the details of each key component below.

\subsection{Boundary Guidance Module} 
\label{boundary}

Several previous works \cite{ding2019boundary,zhang2019net} have demonstrated that boundary information is helpful to improve the performance of computer vision tasks. For example, Ding~\etal\cite{ding2019boundary} propose to learn the boundary as an additional semantic class to enable the network to be aware of the boundary layout for scene segmentation. In the COD task, since camouflaged objects are visually embedded in their background, which makes that the boundary between a camouflaged object and its surrounding background is not sharp. Therefore, it is critical to locate the boundaries of camouflaged objects, in which boundary information provides useful constraints to guide feature extraction during camouflaged object detection. Existing works \cite{ding2019boundary,zhang2019net,fan2020inf} have shown only low-level features preserve sufficient boundary information, thus we carry out BGM on the first low-level layers, \ie, $f_1$ and $f_2$, as shown in Fig.~\ref{fig2}. Specifically, $f_1$ and $f_2$ are fed a $3\times{3}$ convolutional layer, respectively, and then integrated via an addition operation to obtain the fused feature representation. Then, the fused feature is further fed into a $3\times{3}$ convolutional layer to obtain $f_e$, which acts as the edge guidance feature in the decoder path. Moreover, $f_e$ is fed into a $3\times{3}$ convolutional layer to produce the final boundary map, which is upsampled to the same resolution as the original image. Therefore, the produced boundary map and its detection edge map can be measured using the binary cross-entropy loss function, which is given as
\begin{equation}
\begin{split}
\mathcal{L}_{\textup{edge}}=-\sum\nolimits_i[{E}_i^{d}log({E}_i^{p})+(1-{E}_i^{d})log(1-{E}_i^{p})],
\end{split}
\label{eq0-01}
\end{equation}
where ${E}_i^{p}$ denotes the produced boundary map of the $i$-th image, and ${E}_i^{d}$ denotes the boundary ground-truth map. In our model, the Canny edge detection method is used to extract ${E}_i^{d}$. It is worth noting that the early convolutional layers are supervised by the boundary detection loss. Besides, our BGM can provide boundary enhanced representation (\ie, $f_e$) to guide the process of detection in the decoder path. Further, $f_e$ is cascaded to multiple supervisions to enhance the ability of feature representations. We also note that the proposed BGM is different from the Boundary-aware Feature Propagation module in~\cite{ding2019boundary}, we focus on learning the boundary-enhanced representation for preserving the local characteristics and boundary information, which can be incorporated into the decoder network within a layer-wise strategy to improve the COD performance.

\begin{figure}[t]
	\begin{centering}
		\includegraphics[width=0.46\textwidth]{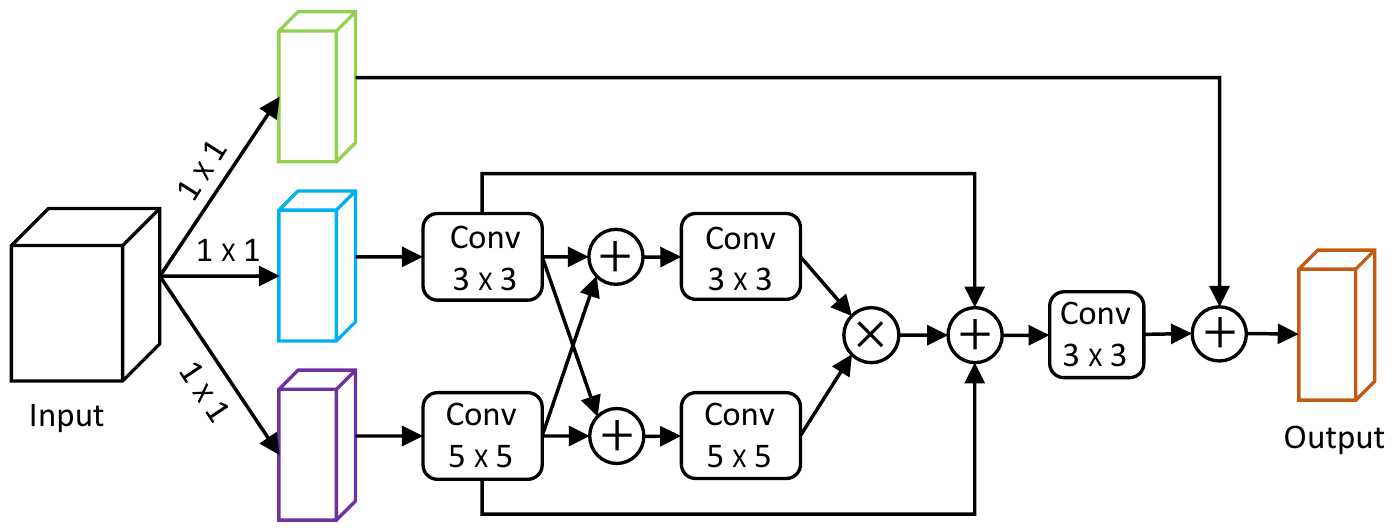}
		\caption{The detailed architecture of the proposed multi-scale feature aggregation module (MFAM).}\vspace{-0.45cm}
		\label{fig3}
	\end{centering}
\end{figure}

\begin{figure*}[t]
	\begin{centering}
		\includegraphics[width=0.97\textwidth]{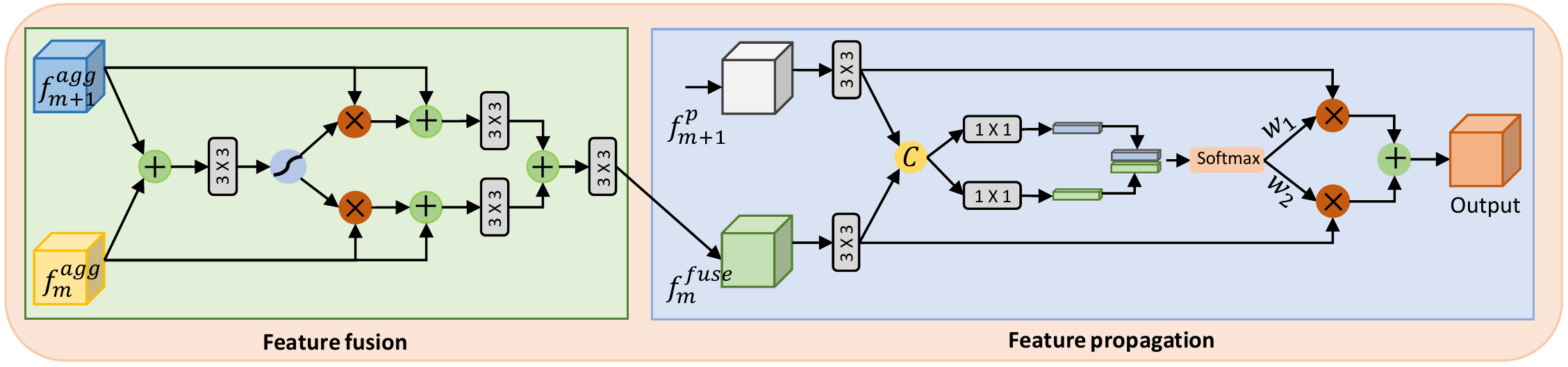}\vspace{-0.1cm}
		\caption{Illustration of the proposed cross-level fusion and propagation module, consisting of two key components: cross-level feature fusion and decoder feature propagation.}
	\end{centering}
\end{figure*}

\subsection{Multi-scale Feature Aggregation Module} 
\label{aggregation}

Scale variation is one of the major challenges in the COD task. Because each convolutional layer only is able to handle a special scale, it is demanded to capture multi-scale information from a single layer to characterize variations of the object's scale. Therefore, we propose a Multi-scale Feature Aggregation Module (MFAM) to aggregate the image features at different scales.

Specifically, we denote the $m$-th level feature as $f_{m}\in\mathbb{R}^{W_m*H_m*C_m}$, where $W_m$, $H_m$, and $C_m$ are the width, height, and channel number, respectively. We then obtain three feature representations (\ie, $f_{m}^1\in\mathbb{R}^{W_m*H_m*C_l}$, $f_{m}^2\in\mathbb{R}^{W_m*H_m*C_l}$, and $f_{m}^3\in\mathbb{R}^{W_m*H_m*C_l}$) by conducting three independent $1\times1$ convolutional layers on $f_{m}$, in which the channel number is reduced to $C_l$ for acceleration. Then we construct a two-stream network using different convolutional kernels. In this case, the information between the two-stream network can be shared with each other for capturing features at different scales. As shown in Fig.~\ref{fig3}, $f_{m}^1$ and $f_{m}^2$ are fed into a $3\times{3}$ convolutional layer and a $5\times{5}$ convolutional layer, respectively, we can obtain $\zeta({Conv_3(f_{m}^1)})$ and $\zeta({Conv_5(f_{m}^2)})$, where $\zeta(\cdot)$ denotes the \emph{ReLU} activation function. Further, the two features are fused and then fed into two convolutional layers with different kernels. The above process can be described as follows:
\begin{equation}
\left\{
\begin{aligned}
f_m^{3\times{3}}=\zeta({Conv_3(\zeta({Conv_3(f_{m}^1)})\oplus{\zeta}({Conv_5(f_{m}^2)}))}),\\ 
f_m^{5\times{5}}=\zeta({Conv_5(\zeta({Conv_3(f_{m}^1)})\oplus{\zeta}({Conv_5(f_{m}^2)}))}),\\
\end{aligned}
\right.
\end{equation}
where $\oplus$ denotes element-wise addition. After that, we can obtain the aggregated feature, \ie, $f_m^{3\times{3}}\otimes f_m^{5\times{5}}$, where $\otimes$ denotes element-wise multiplication. In order to fully exploit feature complementary, the fused multi-scale feature can be formulated by $f_m^{fused}=f_m^{3\times{3}}\otimes f_m^{5\times{5}}+\zeta({Conv_3(f_{m}^1)}+\zeta({Conv_5(f_{m}^2)}$. Moreover, to preserve the original feature information, we also pile the original feature (\ie, $f_m^{3}$) on the fused feature $f_m^{fused}$. Therefore, we can obtain the final multi-scale aggregated feature as
\begin{equation}
\begin{aligned}
f_m^{agg}=\zeta(Conv_3(f_m^{fused})) \oplus f_m^{3}.
\end{aligned}
\end{equation}

It is worth noting that our MFAM introduces different sizes of convolution kernels to adaptively extract features in different scales, and multi-scale features interact with each other to produce more effective and discriminate image information. Then, the aggregated features can be obtained by fusing multi-scale features with a residual connection, which makes that our model can effectively deal with scale variations.

\subsection{Cross-level Fusion and Propagation Module} 
\label{fusion}

Effective fusion of cross-level features by exploiting their correlations often boosts the learning performance. In addition, accurate camouflaged object detection usually relies on the effective features provided by the encoder. However, existing methods directly pass all features to the decoder network, while ignoring the contributions of features from different levels. In this case, the valuable context information can be adequately propagated into the decoder, resulting in unsatisfactory detection results. To lighten these effects, we propose a Cross-level Fusion and Propagation Module (CFPM), which includes two key parts, \ie, cross-feature fusion and feature propagation. 

Specifically, in the cross-level feature fusion part, we take the two cross-level aggregated features $f_{m+1}^{agg}$ and $f_{m}^{agg}$ (here $f_{m+1}^{agg}$ is imposed using an upsampling operation to have the same resolution with $f_{m}^{agg}$) as an example, the two features are first fused by an addition operation. Then, the fused feature is fed into a $3\times{3}$ convolutional layer with a \emph{Sigmoid} activation function, and then we can obtain the normalized feature maps, \ie, $\sigma(Conv_3(f_{m+1}^{agg} \oplus f_{m}^{agg}))\in[0,1]$, where $\sigma(\cdot)$  indicates the \emph{Sigmoid} function. Therefore, the normalized feature maps can be considered as feature-level attention maps to adaptively enhance the feature representation. In this case, the fused feature map is used to enhance the cross-level features to capture their correlation. Besides, to preserve the original information of each feature, a residual connection is adapted to combine the enhanced features with their original features. Therefore, we obtain the enhanced features as follows: 
\begin{equation}
\left\{
\begin{aligned}
&f_{m+1}^{en}=f_{m+1}^{agg}+\sigma(Conv_3(f_{m+1}^{agg}))\otimes f_{m+1}^{agg},\\ 
&f_{m}^{en}=f_{m}^{agg}+\sigma(Conv_3(f_{m+1}^{agg}))\otimes f_{m}^{agg},\\
\end{aligned}
\right.
\end{equation}

Then, we obtain the fused cross-level feature as follows:
\begin{equation}
\begin{aligned}
f_m^{fuse}=Conv_3(Conv_3(f_{m+1}^{en})+Conv_3(f_{m}^{en})).
\end{aligned}
\end{equation}

In the feature propagation part, it is important to propagate the fused features from the encoder to the decoder by combing the output of the previous CFPM. For convenience, we denote the output of the previous CFPM as $f_{m+1}^p\in\mathbb{R}^{C\times{W}\times{H}}$. Firstly, the two features ($f_m^{fuse}$ and $f_{m+1}^p$) are processed through a convolutional layer with $3\times{3}$ kernel size to produce the smooth features (for convenience, we denote them as $p_1$ and $p_2$, respectively). We then concatenate the two feature maps to combine their features at a certain position in space, \ie, $f_{cat}=[p_1,p_2]\in\mathbb{R}^{2C\times{W}\times{H}}$. Further, we carry out a $1\times{1}$ convolution to map the high-dimensional feature (\ie, $f_{cat}$) to two different spatial-wise gates, which are $g_1\in\mathbb{R}^{1\times{W}\times{H}}$ and $g_2\in\mathbb{R}^{1\times{W}\times{H}}$. A softmax function is applied to these two gates, thus we can obtain 
\begin{equation}
\begin{aligned}
w_1^{(i,j)}=\frac{e^{g_{1}^{(i,j)}}}{e^{g_1^{(i,j)}}+e^{g_2^{(i,j)}}};~~~ w_2^{(i,j)}=\frac{e^{g_2^{(i,j)}}}{e^{g_1^{(i,j)}}+e^{g_2^{(i,j)}}},
\end{aligned}
\label{eq005}
\end{equation}
where $g_1^{(i,j)}$ and $g_2^{(i,j)}$ denote the weights assigned for the $(i,j)$-th position in the two feature maps, respectively. Besides, we have $w_1^{(i,j)}+w_2^{(i,j)}=1$. Therefore, the final propagated feature is given by
\begin{equation}
\begin{aligned}
f_m^{p(i,j)}=w_1^{(i,j)}\cdot{p_1^{(i,j)}}+w_2^{(i,j)}\cdot{p_2^{(i,j)}}.
\end{aligned}
\label{eq006}
\end{equation}

So far, we have obtained the fused feature representation $f_m^{p}$ by adaptively combing the features from the encoder layer and decoder path. More importantly, our module can assign weights according to the contributions of the encoder and decoder streams to boost the COD performance. Note that, we only conduct the feature fusion without the feature propagation part in the CFPM when fusing $f_5$ and $f_4$.

\subsection{Overall Loss Function}
\label{loss}

The binary cross-entropy ($\mathcal{L}_{\textup{BCE}}$) is one of the most widely adopted losses in segmentation tasks, however, it ignores the global structure of an image when computing the loss for each pixel independently. Inspired by the success and effectiveness of the standard Intersection-over-Union (IoU) loss and weighted IoU loss in salient object detection \cite{wei2020f3net}, our detection loss function is defined as $\mathcal{L}_{\textup{det}}=\mathcal{L}_{\textup{IoU}}^w+\mathcal{L}_{\textup{BCE}}^w$, where $\mathcal{L}_{\textup{IoU}}^w$ and $\mathcal{L}_{\textup{BCE}}^w$ denote the weighted IoU loss and BCE loss for the global and local restrictions, respectively. It is worth noting that $\mathcal{L}_{\textup{IoU}}^w$ can increase the weights of hard pixels to highlight their importance, and $\mathcal{L}_{\textup{BCE}}^w$ pays more attention to hard pixels rather than treating all pixels equally. Besides, as shown in Fig.~\ref{fig2}, we utilize multiple supervisions for the four side-output maps and the ground-truth map. Here, each map (\ie, $S_i^{up}$) is up-sampled to have the same size as the ground-truth map (\ie, $G$). Therefore, the overall loss function can be formulated as follows:
\begin{equation}
\begin{aligned}
\mathcal{L}_{\textup{total}}=\mathcal{L}_{\textup{edge}}+\sum_{i=1}^4\mathcal{L}_{\textup{det}}(G,S_i^{up}).
\end{aligned}
\label{eq13}
\end{equation}

\begin{table*}[t!]
  \centering
  \renewcommand{\arraystretch}{0.98}
  \renewcommand{\tabcolsep}{1.6mm}
  \caption{
  Quantitative comparison our model with 20 state-of-the-art models using four evaluation metrics (\ie, $S_{\alpha}$ \cite{fan2017structure}, mean $F_{\beta}$ \cite{achanta2009frequency}, mean $E_{\phi}$\cite{Fan2018Enhanced}, and $\mathcal{M}$ \cite{perazzi2012saliency}). ``$\uparrow$`` \& ``$\downarrow$'' indicate that larger or smaller is better. The best results are highlighted in \textbf{Bold} fonts.} \vspace{-0.25cm}
  \begin{tabular}{r||c||cccc|cccc|cccc}
  \whline{1pt} 
    
    \multirow{2}*{Methods}
    &\multirow{2}*{\#Param (M)}
    &\multicolumn{4}{c|}{CHAMELEON}
    &\multicolumn{4}{c|}{CAMO-Test}
    &\multicolumn{4}{c}{COD10K-Test}\\
    \cline{3-14} 

    &
    &$S_{\alpha}\uparrow$ ~  &$F_{\beta}\uparrow$ ~  &$E_{\phi}\uparrow$ ~  &$\mathcal{M}\downarrow$  
    &$S_{\alpha}\uparrow$ ~  &$F_{\beta}\uparrow$ ~  &$E_{\phi}\uparrow$ ~  &$\mathcal{M}\downarrow$ 
    &$S_{\alpha}\uparrow$ ~  &$F_{\beta}\uparrow$ ~  &$E_{\phi}\uparrow$ ~  &$\mathcal{M}\downarrow$ \\

    \whline{1pt} 
    FPN~\cite{fpn}
    &-
    & 0.794  & 0.648  & 0.783  & 0.075 
    & 0.684  & 0.545  & 0.677  & 0.131 
    & 0.697  & 0.481  & 0.691  & 0.075 \\ 

    MaskRCNN~\cite{maskrcnn}
    &- & 0.643  & 0.610  & 0.778  & 0.099 
    & 0.574  & 0.520  & 0.715  & 0.151 
    & 0.613  & 0.469  & 0.748  & 0.081 \\ 
    
    PSPNet~\cite{pspnet} 
    & 71.64 & 0.773 & 0.630 & 0.758 & 0.085 
    & 0.663 & 0.520 & 0.659 & 0.139 
    & 0.678 & 0.458 & 0.680 & 0.080 \\ 
    
    UNet++~\cite{zhou2018unet++} 
    & 36.63 & 0.695 & 0.557 & 0.762 & 0.094 
    & 0.599 & 0.461 & 0.653 & 0.149 
    & 0.623 & 0.409 & 0.673 & 0.086 \\ 
    
    PiCANet~\cite{picanet} 
    & 32.85 & 0.769 & 0.615 & 0.749 & 0.085 
    & 0.609 & 0.419 & 0.584 & 0.156 
    & 0.649 & 0.411 & 0.643 & 0.090 \\ 
    

    MSRCNN~\cite{masksrcnn} 
    &-& 0.637 & 0.505 & 0.686 & 0.091 
    & 0.617 & 0.527 & 0.669 & 0.133 
    & 0.641 & 0.478 & 0.706 & 0.073 \\
    
    BASNet~\cite{basnet} 
    & 87.06 & 0.687 & 0.528 & 0.721 & 0.118 
    & 0.618 & 0.475 & 0.661 & 0.159 
    & 0.634 & 0.417 & 0.678 & 0.105 \\
    
    PFANet~\cite{pfanet} 
    & 16.38 & 0.679 & 0.648 & 0.378 & 0.144 
    & 0.659 & 0.622 & 0.391 & 0.172 
    & 0.636 & 0.618 & 0.286 & 0.128 \\
    
    CPD~\cite{cpd} 
    & 29.23 & 0.857 & 0.771 & 0.874 & 0.048 
    & 0.716 & 0.618 & 0.723 & 0.113 
    & 0.750 & 0.595 & 0.776 & 0.053 \\ 
    
    HTC~\cite{htc} 
    &-& 0.517 & 0.236 & 0.489 & 0.129 
    & 0.476 & 0.206 & 0.442 & 0.172 
    & 0.548 & 0.253 & 0.520 & 0.088 \\ 
    
    PoolNet~\cite{liu2019simple} 
    & 53.63 & 0.776 & 0.632 & 0.779 & 0.081 
    & 0.703 & 0.563 & 0.699 & 0.129 
    & 0.705 & 0.500 & 0.713 & 0.074 \\ 
    
    EGNet~\cite{egnet} 
    & 108.07 & 0.750 & 0.645 & 0.764 & 0.075 
    & 0.662 & 0.567 & 0.683 & 0.125 
    & 0.733 & 0.583 & 0.761 & 0.055 \\ 
    
    
    
    GateNet~\cite{zhao2020suppress} 
    & 128.63 & 0.870 & 0.775 & 0.873 & 0.045 
    & 0.773 & 0.684 & 0.771 & 0.093 
    & 0.791 & 0.643 & 0.797 & 0.047 \\ 
    
    MINet~\cite{pang2020multi} 
    & 162.38 & 0.868 & 0.767 & 0.869 & 0.048 
    & 0.773 & 0.678 & 0.771 & 0.100 
    & 0.786 & 0.639 & 0.803 & 0.052 \\ 
    
    
    PraNet~\cite{fan2020pranet} 
    & 32.55 & 0.860 & 0.789 & 0.907 & 0.050 
    & 0.769 & 0.710 & 0.825 & 0.094 
    & 0.789 & 0.671 & 0.861 & 0.045 \\ 

    SINet~\cite{fan2020camouflaged} 
    & 48.95 & 0.869 & 0.790 & 0.891 & 0.044 
    & 0.751 & 0.675 & 0.771 & 0.100 
    & 0.771 & 0.634 & 0.807 & 0.051 \\ 

    POCINet~\cite{liu2021integrating} 
    &-& 0.866 & 0.807 & 0.905 & 0.042
    & 0.702 & 0.629 & 0.731 & 0.110 
    & 0.751 & 0.628 & 0.810 & 0.051 \\ 

    DNTDF~\cite{fang2022densely} 
    &28.85& 0.864 & 0.773 & 0.881 & 0.046
    & 0.772 & 0.682 & 0.784 & 0.097 
    & 0.781 & 0.629 & 0.800 & 0.049 \\ 
    
    
    LSR~\cite{lv2021} 
    &-& 0.890 & 0.841 & 0.935 & 0.031
    & 0.787 & 0.744 & 0.838 & 0.080 
    & 0.804 & 0.715 & 0.880 & 0.037 \\ 
    
    PFNet~\cite{mei2021camouflaged} 
    & 46.50 & 0.882 & 0.828 & 0.931 & 0.033 
    & 0.782 & 0.746 & 0.842 & 0.085 
    & 0.800 & 0.701 & 0.877 & 0.040 \\

    \whline{1pt} 
    
    Ours
    & 29.52 & \textbf{0.893} & \textbf{0.842} & \textbf{0.940} & \textbf{0.028} 
    & \textbf{0.815} & \textbf{0.776} & \textbf{0.865} & \textbf{0.076} 
    & \textbf{0.822} & \textbf{0.731} & \textbf{0.888} & \textbf{0.036} \\

   \whline{1pt} 
  \end{tabular}\label{tab1}
\end{table*}

\section{Experiments}
\label{Experiments}

In this section, we first present the experimental settings, including the datasets, evaluation metrics, and implementation details. Then we present the comparison results between our model and other state-of-the-art methods, and we conduct ablation studies to validate the effectiveness of each key
component. Finally, we extend the application of the proposed model to polyp segmentation.

\subsection{Experimental Setup}

\textbf{Datasets}. We conduct experiments on three public datasets for camouflaged object detection. 
$\bullet$ CHAMELEON~\cite{fan2020camouflaged} is collected via the Google search engine with the keyword ``camouflage animals", containing $76$ camouflaged images, which are all used for testing. $\bullet$ CAMO~\cite{le2019anabranch} has $1,250$ images with 8 categories, of which $1,000$ images are for training and the remaining $250$ ones are for testing. $\bullet$ COD10K~\cite{fan2020camouflaged} is currently the largest camouflaged object dataset with high-quality pixel-level annotations. There is a total of $5,066$ camouflaged images in this dataset, where $3,040$ images are for training and $2,026$ ones are for testing. It is also divided into $5$ super-classes and $69$ sub-classes.

\textbf{Evaluation Metrics}. To comprehensively compare our proposed model with other state-of-the-art methods, we adopt five popular metrics to evaluate the COD performance. The details of each metric are provided as follows.

1) {\textbf{PR Curve}}. Given a saliency map $S$, we can convert it to a binary mask $M$, and then compute the \emph{precision}
and \emph{recall} by comparing $M$ with ground-truth $G$:
\begin{equation}
\textup{Precision}=\frac{|M\cap{G}|}{|M|},~ \textup{Recall}=\frac{|M\cap{G}|}{|G|}.
\end{equation}

Then, we adopt a popular strategy to partition the saliency map $S$ using a set of thresholds (\ie, from $0$ to $255$). For each threshold, we first calculate a pair of recall and precision scores, and then combine them to obtain a PR curve that describes the performance of the model at the different thresholds.

2) \emph{Structure Measure ($S_{\alpha}$)}\cite{fan2017structure}: It is proposed to assess the structural similarity between the regional perception ($S_r$) and object perception ($S_o$), which is defined by
\begin{equation}
\begin{aligned}
S_{\alpha}=\alpha \times S_{o}+\left(1-\alpha\right) \times S_{r},
\end{aligned}
\end{equation}
where $ \alpha \in \left[ 0,1\right]$ is a trade-off parameter and it is set to $0.5$ as default \cite{fan2017structure}.

3) \emph{Enhanced-alignment Measure ($E_{\phi}$)} \cite{Fan2018Enhanced}: It is used to capture image-level statistics and their local pixel matching information, which is defined by
\begin{equation}
\begin{aligned}
E_{\phi}=\frac{1}{W \times H}\sum_{i=1}^{W}\sum_{j=1}^{H}\phi\left(S(x,y), G(x,y)\right),
\end{aligned}
\end{equation}
where $W$ and $H$ denote the width and height of ground-truth $G$, and $(x,y)$ is the coordinate of each pixel in $G$. Symbol $\phi$ is the enhanced alignment matrix. We obtain a set of $E_{\phi}$ by converting the prediction $S$ into a binary mask with a threshold in the range of $[0, 255]$. In our experiments, we report the mean of $E_{\phi}$ values over all the thresholds.

4) \emph{F-measure} ($F_{\beta}$) \cite{achanta2009frequency}: It is used to comprehensively consider both precision and recall, and we can obtain the weighted harmonic mean by 
\begin{equation}
\begin{aligned}
F_{\beta}=\left(1+\beta ^2\right)\frac{\textup{Precision} \times \textup{Recall}}{\beta^{2}\textup{Precision}+\textup{Recall}},
\end{aligned}
\end{equation}
where $\textup{Precision}=\frac{|M\cap{G}|}{|M|}$ and $\textup{Recall}=\frac{|M\cap{G}|}{|G|}$. Besides, $\beta^2$ is set to $0.3$ to emphasize the precision \cite{achanta2009frequency}. We use different fixed $[0,255]$ thresholds to compute the $F$-measure. This yields a set of $F$-measure values for which we report the maximal $F_{\beta}$ in our experiments. 

5) \emph{Mean Absolute Error ($\mathcal{M}$)} \cite{perazzi2012saliency}: It is adopted to compute the average pixel-level relative error between the ground truth and normalized prediction, which is defined by
\begin{equation}
\mathcal{M}=\frac{1}{W \times H}\sum_{i=1}^{W}\sum_{j=1}^{H}\left|S\left(i,j\right)-G\left(i,j\right) \right|,
\end{equation}
where $G$ and $S$ denote the ground truth and normalized prediction (it is normalized to $[0,1]$).\\

\begin{figure*}[t]
\centering
	\includegraphics[width=0.95\textwidth]{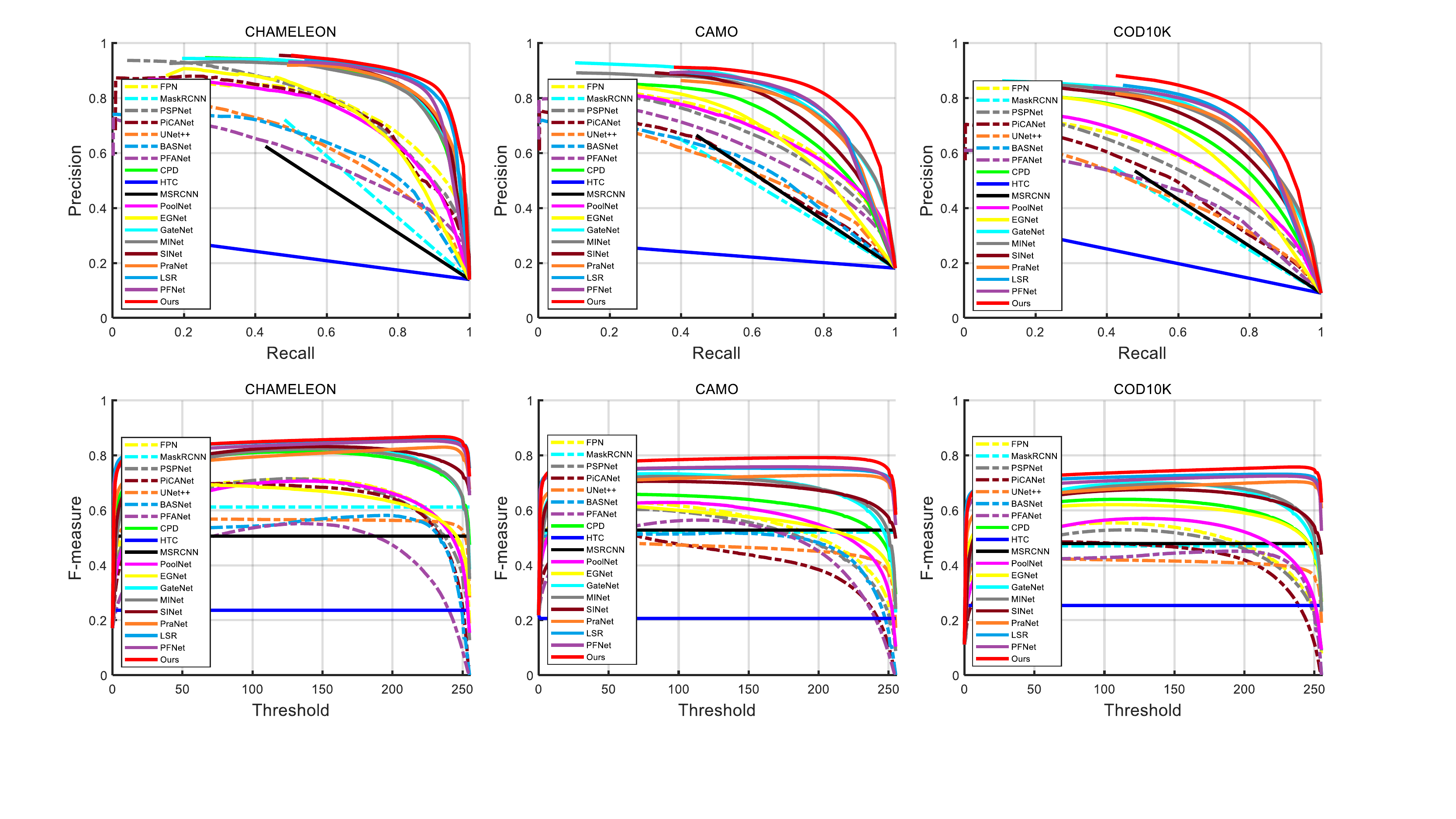}\vspace{-0.3cm}
\caption{Precision-recall (top) and F-measure (bottom) curves on the three camouflaged object datasets.}  
    \label{fig061}
\end{figure*}

\begin{table*}[t!]
  \centering
  \renewcommand{\arraystretch}{1.0}
  \renewcommand{\tabcolsep}{2.0mm}
  \caption{
  Quantitative results on four super-classes (\ie, Flying, Terrestrial, Aquatic, and Amphibian) of the COD10K dataset using four evaluation metrics (\ie, $S_{\alpha}$ \cite{fan2017structure}, mean $F_{\beta}$ \cite{achanta2009frequency}, mean $E_{\phi}$\cite{Fan2018Enhanced}, and $\mathcal{M}$ \cite{perazzi2012saliency}). 
  ``$\uparrow$`` \& ``$\downarrow$'' indicate that larger or smaller is better. The best results are highlighted in \textbf{Bold} fonts. 
  } \vspace{-0.15cm}
  \scriptsize
  \begin{tabular}{r||cccc|cccc|cccc|cccc}
  \whline{1pt} 
    
    \multirow{2}*{Methods}
    &\multicolumn{4}{c|}{Flying (714 images)}
    &\multicolumn{4}{c|}{Terrestrial (699 images)}
    &\multicolumn{4}{c|}{Aquatic (474 images)}
    &\multicolumn{4}{c}{Amphibian (124 images)}\\
    \cline{2-17} 

   
    &$S_{\alpha}\uparrow$   &$F_{\beta}\uparrow$   &$E_{\phi}\uparrow$   &$\mathcal{M}\downarrow$  
    &$S_{\alpha}\uparrow$   &$F_{\beta}\uparrow$   &$E_{\phi}\uparrow$   &$\mathcal{M}\downarrow$ 
    &$S_{\alpha}\uparrow$   &$F_{\beta}\uparrow$   &$E_{\phi}\uparrow$   &$\mathcal{M}\downarrow$  
    &$S_{\alpha}\uparrow$   &$F_{\beta}\uparrow$   &$E_{\phi}\uparrow$   &$\mathcal{M}\downarrow$ \\

    \whline{1pt} 
    
    FPN~\cite{fpn}
    & 0.726  & 0.510  & 0.714  & 0.061 
    & 0.669  & 0.418  & 0.661  & 0.071 
    & 0.684  & 0.509  & 0.687  & 0.103
    & 0.744  & 0.569  & 0.743  & 0.065 \\ 

    MaskRCNN~\cite{maskrcnn}
    & 0.645  & 0.518  & 0.765  & 0.063 
    & 0.608  & 0.440  & 0.747  & 0.070 
    & 0.560  & 0.417  & 0.719  & 0.123
    & 0.665  & 0.552  & 0.782  & 0.081 \\ 
    
    PSPNet~\cite{pspnet} 
    & 0.700 & 0.477 & 0.692 & 0.067 
    & 0.658 & 0.407 & 0.666 & 0.074 
    & 0.659 & 0.478 & 0.670 & 0.111
    & 0.736 & 0.556 & 0.733 & 0.072 \\ 

    UNet++~\cite{zhou2018unet++} 
    & 0.659 & 0.455 & 0.708 & 0.068 
    & 0.593 & 0.340 & 0.637 & 0.081
    & 0.599 & 0.418 & 0.660 & 0.121
    & 0.677 & 0.496 & 0.725 & 0.079 \\ 
    
    PiCANet~\cite{picanet} 
    & 0.677 & 0.440 & 0.663 & 0.076 
    & 0.625 & 0.359 & 0.628 & 0.084
    & 0.629 & 0.423 & 0.623 & 0.120
    & 0.704 & 0.494 & 0.689 & 0.086 \\ 

    MSRCNN~\cite{masksrcnn} 
    & 0.675 & 0.522 & 0.742 & 0.058 
    & 0.611 & 0.417 & 0.671 & 0.070 
    & 0.614 & 0.464 & 0.685 & 0.107
    & 0.722 & 0.613 & 0.784 & 0.055 \\ 

    BASNet~\cite{basnet} 
    & 0.664 & 0.454 & 0.710 & 0.086 
    & 0.601 & 0.350 & 0.645 & 0.109
    & 0.620 & 0.431 & 0.666 & 0.134
    & 0.708 & 0.535 & 0.741 & 0.087 \\
     
    PFANet~\cite{pfanet} 
    & 0.657 & 0.393 & 0.632 & 0.113 
    & 0.609 & 0.323 & 0.600 & 0.123 
    & 0.629 & 0.404 & 0.614 & 0.162
    & 0.690 & 0.460 & 0.661 & 0.119 \\
    
    CPD~\cite{cpd} 
    & 0.777 & 0.624 & 0.792 & 0.041 
    & 0.714 & 0.526 & 0.747 & 0.053 
    & 0.746 & 0.628 & 0.779 & 0.075
    & 0.816 & 0.700 & 0.847 & 0.041 \\
    
    HTC~\cite{htc} 
    & 0.582 & 0.308 & 0.558 & 0.070 
    & 0.530 & 0.196 & 0.484 & 0.078 
    & 0.507 & 0.223 & 0.494 & 0.129
    & 0.606 & 0.365 & 0.596 & 0.088 \\ 
    
    PoolNet~\cite{liu2019simple} 
    & 0.733 & 0.534 & 0.734 & 0.062 
    & 0.677 & 0.442 & 0.688 & 0.071 
    & 0.689 & 0.507 & 0.705 & 0.102
    & 0.767 & 0.598 & 0.769 & 0.064 \\ 
    
    EGNet~\cite{egnet} 
    & 0.771 & 0.630 & 0.795 & 0.040 
    & 0.711 & 0.531 & 0.738 & 0.049 
    & 0.693 & 0.568 & 0.730 & 0.088
    & 0.787 & 0.669 & 0.823 & 0.048 \\ 

    GateNet~\cite{zhao2020suppress} 
    & 0.819 & 0.675 & 0.823 & 0.036 
    & 0.760 & 0.579 & 0.758 & 0.048 
    & 0.784 & 0.672 & 0.804 & 0.064
    & 0.838 & 0.714 & 0.841 & 0.040 \\

    MINet~\cite{pang2020multi} 
    & 0.807 & 0.718 & 0.886 & 0.030 
    & 0.742 & 0.617 & 0.830 & 0.042 
    & 0.767 & 0.703 & 0.843 & 0.060
    & 0.827 & 0.756 & 0.897 & 0.034 \\ 
    
    PraNet~\cite{fan2020pranet} 
    & 0.819 & 0.707 & 0.888 & 0.033 
    & 0.756 & 0.607 & 0.835 & 0.046 
    & 0.781 & 0.692 & 0.848 & 0.065
    & 0.842 & 0.750 & 0.905 & 0.035 \\ 

    SINet~\cite{fan2020camouflaged} 
    & 0.798 & 0.663 & 0.828 & 0.040 
    & 0.743 & 0.578 & 0.778 & 0.050
    & 0.758 & 0.650 & 0.803 & 0.073 
    & 0.827 & 0.724 & 0.866 & 0.042 \\

    LSR~\cite{lv2021} 
    & 0.830 & 0.745 & {0.906} & 0.027 
    & 0.772 & 0.656 & 0.855 & 0.038 
    & 0.803 & 0.740 & 0.875 & 0.053
    & 0.846 & 0.783 & 0.906 & \textbf{0.030} \\ 
     
    PFNet~\cite{mei2021camouflaged} 
    & 0.824 & 0.729 & 0.903 & 0.030
    & 0.773 & 0.648 & 0.855 & 0.041 
    & 0.793 & 0.722 & 0.868 & 0.055
    & 0.848 & 0.773 & 0.911 & 0.031 \\

    \whline{1pt} 
    
    Ours
    & \textbf{0.845} & \textbf{0.760} & \textbf{0.906} & \textbf{0.025} 
    & \textbf{0.795} & \textbf{0.678} & \textbf{0.868} & \textbf{0.037} 
    & \textbf{0.821} & \textbf{0.757} & \textbf{0.887} & \textbf{0.049}
    & \textbf{0.854} & \textbf{0.783} & \textbf{0.914} & 0.032 \\

   \whline{1pt} 
  \end{tabular}\label{tab2}
\end{table*}

\begin{figure*}[t]
\centering
\begin{overpic}[width=1.0\linewidth]{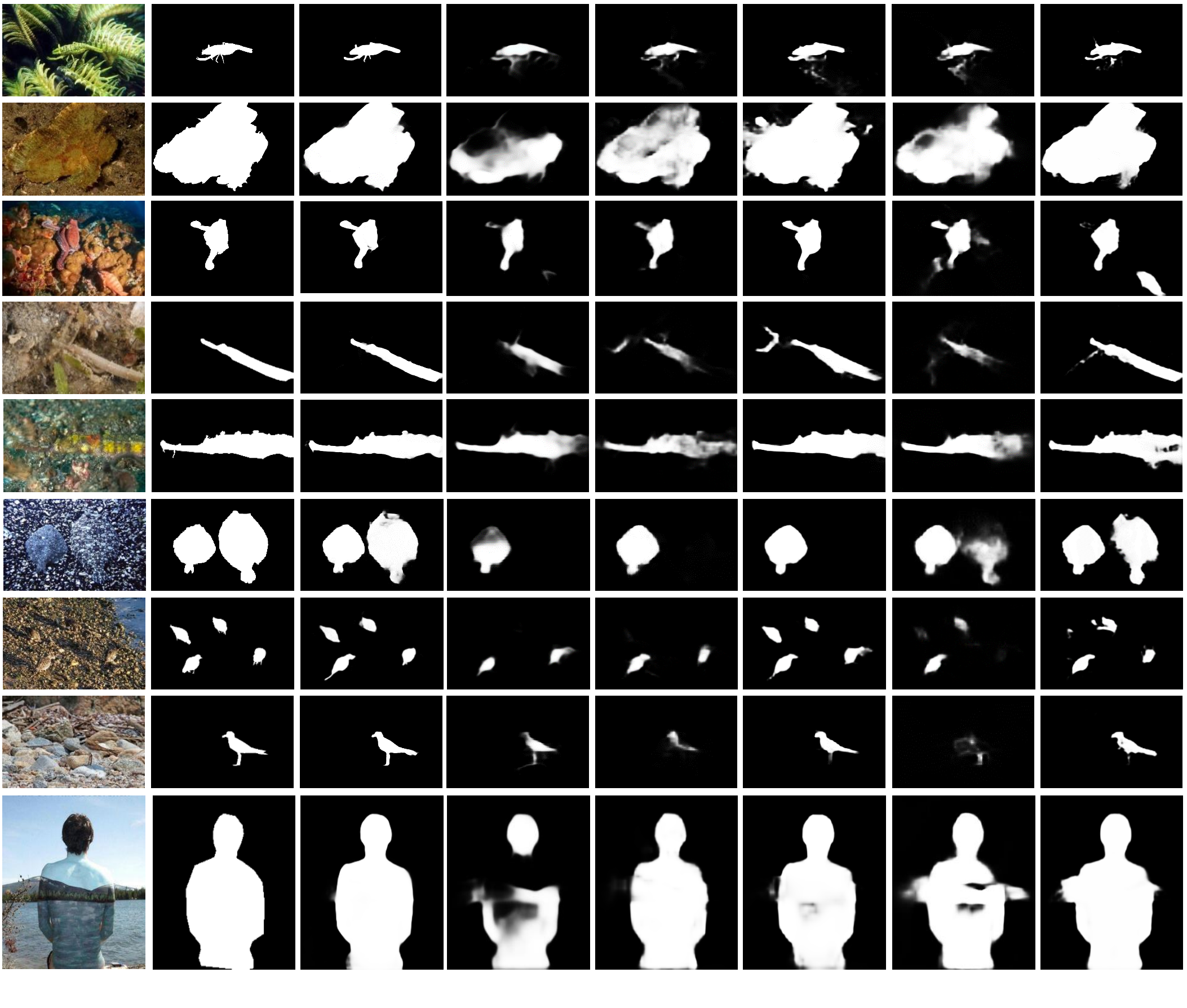}
\put(6, .8){\footnotesize RGB}  
\put(18,.8){\footnotesize GT}  
\put(30,.8){\footnotesize Ours}  
\put(42,.8){\footnotesize EGNet}  
\put(56,.8){\footnotesize CPD}  
\put(67,.8){\footnotesize PraNet}  
\put(80,.8){\footnotesize SINet}  
\put(92,.8){\footnotesize PFNet}  
\end{overpic}\vspace{-0.25cm}

\caption{Qualitative visual comparison of our model versus five state-of-the-art methods (\ie, EGNet \cite{egnet}, CPD~\cite{cpd}, PraNet~\cite{fan2020pranet}, SINet~\cite{fan2020camouflaged}, and PFNet~\cite{mei2021camouflaged}).}  \vspace {-0.25cm}
    \label{fig06}

\end{figure*}

\textbf{Implementation Details}. Our model is implemented in PyTorch and trained on one NVIDIA Tesla P40 GPU with 24 GB memory. The backbone network (Res2Net-50 \cite{pami20Res2net}) is used, which has been pre-trained on ImageNet \cite{russakovsky2015imagenet}. We adopt the Adam algorithm to optimize the proposed model. The initial learning rate is set to $1e-4$ and is divided by $10$ every $30$ epochs. We adopt additional data augmentation strategies including random flipping, crop, and rotation using different scaling ratios, \ie, $\{0.75, 1, 1.25\}$. The input images are resized to $352\times{352}$. The batch size is set to $20$ and the model is trained over $200$ epochs. Following the training setting in~\cite{fan2020camouflaged}, we utilize the default training sets including CAMO and COD10K datasets. Then, we evaluate the proposed model and all compared methods on the whole CHAMELEON dataset and the test sets of CAMO and COD10K datasets. Besides, during the testing stage, the test images are resized to $352\times{352}$ and then fed into the model to obtain prediction maps. Finally, the prediction maps can be rescaled to the original size to achieve the final evaluation.

\subsection{Comparison with State-of-the-art Methods}
\subsubsection{Comparison Methods}

To validate the effectiveness of the proposed COD method, we compare it with 20 state-of-the-art methods, including FPN~\cite{fpn}, MaskRCNN~\cite{maskrcnn}, PSPNet~\cite{pspnet}, UNet++~\cite{zhou2018unet++}, PiCANet~\cite{picanet}, MSRCNN~\cite{masksrcnn}, BASNet~\cite{basnet}, PFANet~\cite{pfanet}, CPD~\cite{cpd}, HTC~\cite{htc}, PoolNet~\cite{liu2019simple}, EGNet~\cite{egnet}, GateNet~\cite{zhao2020suppress}, MINet~\cite{pang2020multi}, PraNet~\cite{fan2020pranet}, SINet~\cite{fan2020camouflaged}, POCINet~\cite{liu2021integrating}, DNTD~\cite{fang2022densely}, LSR~\cite{lv2021}, and PFNet~\cite{mei2021camouflaged}. For GateNet, MINet, and DNTDF, we retrained the three models with released codes. For other all compared methods, we collected the prediction maps from \cite{fan2020camouflaged}. Besides, we evaluate all the prediction maps using the same code.

\subsubsection{Quantitative Comparison}

Table~\ref{tab1} shows the quantitative comparison between our model and 20 state-of-the-art methods by four widely used evaluation metrics. On the CHAMELEON dataset, from the results, it can be seen that our method outperforms all the state-of-the-art methods in all evaluation metrics. LSR and PFNet achieve relatively better COD performance than other comparison methods. Besides, some methods (\eg, EGNet~\cite{egnet} and PFANet~\cite{pfanet}) also utilize auxiliary edge or boundary information and still fail to locate camouflaged objects, while our model can effectively locate them and achieve the best performance. This is because our model can fully capture the multi-scale information to deal with the objects' scale variations, and the fused features cross-level features from the encoder can be propagated to the decoder for providing much useful context information to improve the COD performance. On the CAMO dataset, our model consistently obtains the best performance, further demonstrating its robustness in locating camouflaged objects under challenging factors. Compared with PFNet~\cite{mei2021camouflaged}, our model significantly improves $S_{\alpha}$ by $4.2\%$, $F_{\beta}$ by $4.0\%$, and $E_{\phi}$ by $2.7\%$. Besides, our model significantly improves $S_{\alpha}$ by $6.0\%$, $F_{\beta}$ by $9.3\%$, and $E_{\phi}$ by $4.9\%$ when compared with PraNet~\cite{pspnet}. On the largest COD10K dataset, it can be again observed that our model is consistently better than other compared COD methods. This is because our model can provide boundary-enhanced features to help locate the boundaries of camouflaged objects, resulting in accurate detection of the camouflaged object. 

In addition to the overall quantitative comparisons using the above four evaluation metrics, we show PR and F-measure curves in Fig.~\ref{fig061}. From the results, we can see that our model achieves the best results compared to other COD methods. 

Moreover, to investigate the complexity of the proposed model, we list the number of parameters (\#Param) of different COD methods in Table~\ref{tab1}. From the results, it can be observed that our model is with minimal parameters in comparison with two representative COD methods (\ie, SINet~\cite{fan2020camouflaged} and PFNet~\cite{mei2021camouflaged}).

\subsubsection{Qualitative Comparison}

Fig.~\ref{fig06} shows the detection results of our model and five comparison COD methods. From the visual results in Fig.~\ref{fig06}, our model achieves better visual results by detecting and segmenting more accurate and complete camouflaged objects. Specifically, in the $1^{st}$ and $2^{nd}$  rows, it can be seen that our model can effectively handle size variations, while EGNet, CPD, PraNet, and SINet suffer from inaccurate segmentation results. In the $3^{rd}$ and $4^{th}$ rows, camouflaged objects have a similar texture to the background, which brings a serious challenge to identify them from a similar background. In this case, our model performs better and accurately locates camouflaged objects. In the $5^{th}$ row, the boundary between the object and background is not sharp, while our method still accurately detects the camouflaged object with rich details. Among the five comparison methods, PraNet obtains relatively better performance than other compared methods (see the $5^{th}$ row). In the $6^{th}$ and $7^{th}$ rows, it is challenging to detect multiple camouflaged objects. It can be observed that our model can effectively detect multiple camouflaged objects while some methods fail to locate them. In the $8^{th}$ and $9^{th}$ rows, we can that the objects are visually embedded in their background, thus it is very challenging for COD methods to identify them. In this case, our model detects camouflage objects more accurately than other compared methods. Overall, the results prove that our model can achieve good performance in detecting camouflage objects under different challenging factors.

\subsubsection{Super-class Performance Comparison}

To further verify the effectiveness of the proposed COD model, we report the
quantitative  super-class results in Table~\ref{tab2}. On the three super-classes, \ie, ``Flyingm" ``Terrestrial", and ``Aquatic'', our model obtains the best performance in the terms of four evaluation metrics. Specifically, compared with PraNet, the improvements are $3.2\%$, $12.6\%$, $10.1\%$ in terms of $S_{\alpha}$, $F_{\beta}$, and $E_{\phi}$ on the ``Flying" class.  Compared with SINet, the improvements are $2.6\%$ and $4.3\%$ in terms of $S_{\alpha}$ and $F_{\beta}$, respectively. Our model achieves $4.6\%$, $4.7\%$, and $1.9\%$ improvements in the term of $S_{\alpha}$ over PraNet on the ``Terrestrial", ``Aquatic'', and ``Amphibian" classes, respectively. Besides, our model achieves $17.1\%$, $12.7\%$, and $9.7\%$ improvements in the term of $F_{\beta}$ over PraNet on the ``Terrestrial", `Aquatic'', and ``Amphibian" classes, respectively. Overall, the proposed \ours~achieves satisfactory performance under different super-class conditions.

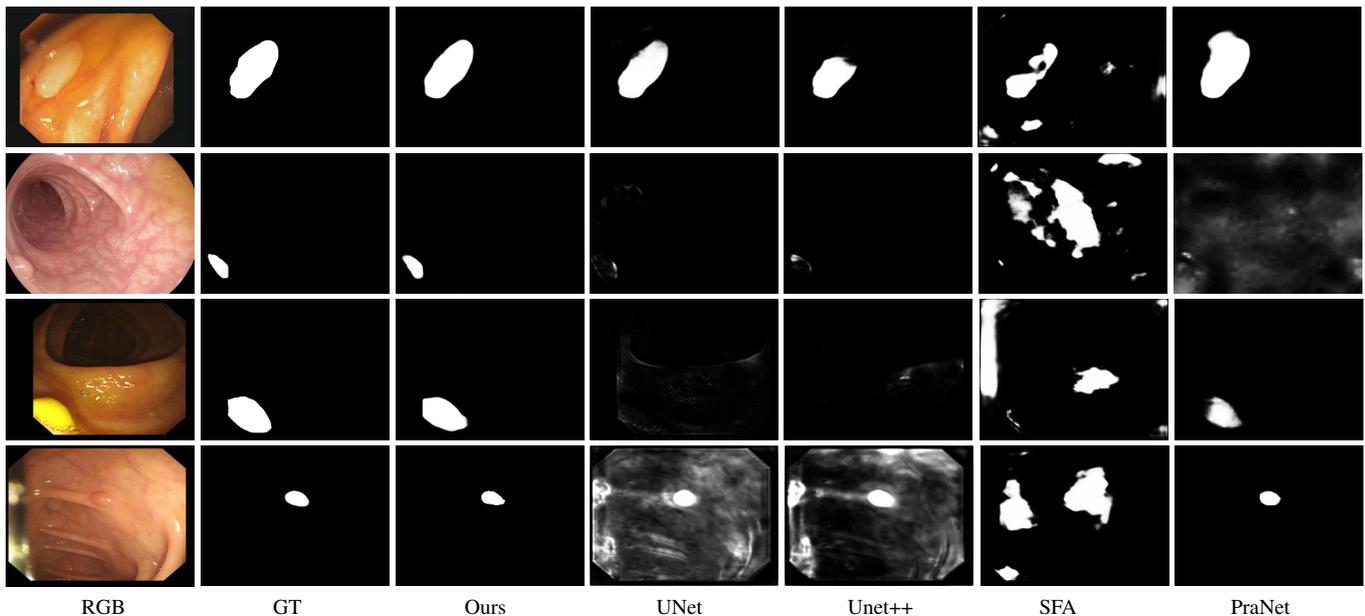
\begin{figure*}[t]
\centering
\begin{overpic}[width=1.0\linewidth]{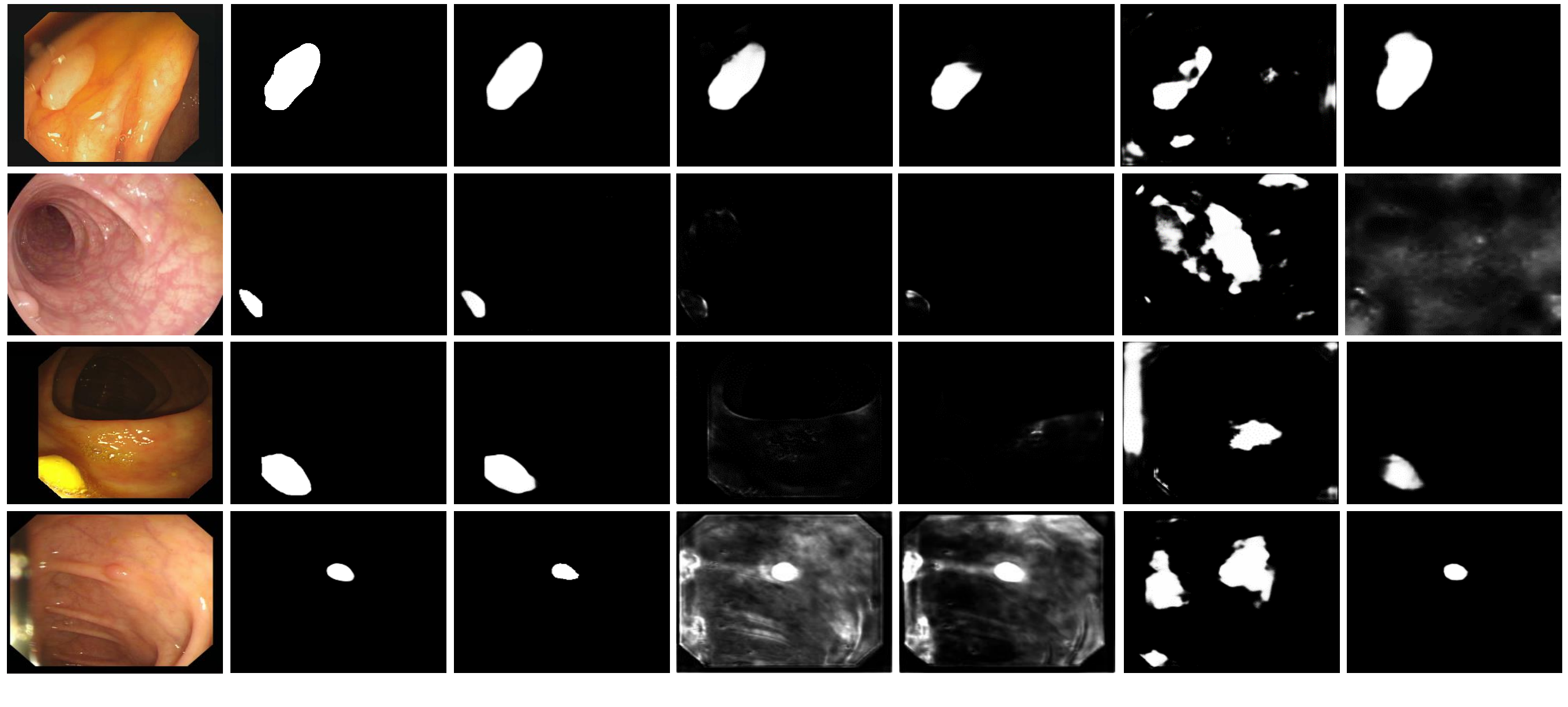}
\put(6, 0.1){\footnotesize RGB}  
\put(20,0.1){\footnotesize GT}  
\put(34,0.1){\footnotesize Ours}  
\put(48,0.1){\footnotesize UNet}  
\put(62,0.1){\footnotesize Unet++}  
\put(76,0.1){\footnotesize SFA}  
\put(90,0.1){\footnotesize PraNet}  
\end{overpic}
\caption{Qualitative visual comparison of different polyp segmentation methods.} 
\vspace {-0.25cm}
 \label{fig07}

\end{figure*}

\subsection{Ablation Study}

To investigate the effectiveness of different key components in the proposed model, we carry out ablation studies by removing or replacing them from our full model. We first remove all key components. Specifically, we first provide the COD results using two baseline methods, \ie, FPN~\cite{fpn} and Attention U-Net~\cite{oktay2018attention} (denoted as ``A1" and ``A2"). We then adopt a $3\times{3}$ convolution to replace the convolution for reducing the channel size, and utilize a simple concatenation followed by a $3\times{3}$ convolution instead of CPFM (this experiment is denoted as ``A3"). In the ``B1" experiment, we add the MFAM, while other key components are removed and the related experimental settings are similar to ``A3". In addition, we adopt the Inception module~\cite{szegedy2015going} instead of MFAM, and this experiment is denoted as ``B2". In the ``C1" experiment, we further add the cross-level feature fusion part of the CFPM, while other key components are removed. In the ``C2" experiment, we utilize the full CFPM, while only BGM is removed. Moreover, in the ``D" experiment, we adopt the Attention Gate in~\cite{oktay2018attention} instead of the proposed CFPM. Finally, we show the experiment of our full model in the ``E" experiment. Experimental results of ablation studies for different key components in our model are shown in Table~\ref{tab04}.

\textbf{Effectiveness of Baseline Model}. As shown in Table~\ref{tab04}, comparing our basic framework (``A3") with other baselines (``A1" and ``A2"), it can be observed that our basic framework performs better than FPN and Attention U-Net, which indicates the effectiveness of our basic framework in detecting camouflaged objects.

\begin{table}[t!]
	\scriptsize
	\centering
	 \renewcommand{\arraystretch}{1.0}
     \renewcommand{\tabcolsep}{1.0mm}
	\caption{Ablation studies for different baseline methods and key components of our model}.\vspace{-0.6cm}
	\begin{center}
	\begin{tabular}{c|c|ccc|cc|cc|c|c} 
	 \whline{1pt} 
		Datasets &{Metrics}                   & A1  & A2    & A3    & B1    & B2   &C1 & C2     & D     & E \\

	 \whline{1pt} 
		\multirow{4}{*}{{{CHAMELEON}}} 
		& $ S_{\alpha}\uparrow $    & .668 & .794 & .868  & .882   & .874 & .887  & .888 & .872  & .893   \\
		& $ E_{\phi}\uparrow   $    & .611 & .783 & .922  & .928   & .928 & .945  & .933 & .924  & .940   \\
		& $\mathcal{M}\downarrow$   & .152 & .075 & .038  & .035   & .034 & .030  & .032 & .035 & .028   \\

		\hline
		\multirow{4}{*}{{{CAMO}}}  
		& $ S_{\alpha}\uparrow $    & .629 & .684 & .798  & .802   & .799 & .800  & .811 & .808  & .815   \\
		& $ E_{\phi}\uparrow   $    & .577 & .677 & .845  & .852   & .846 & .855  & .864 & .852  & .865   \\
		& $\mathcal{M}\downarrow$   & .191 & .131 & .084  & .083   & .083 & .078  & .078 & .079  & .076   \\
		
		\hline
		\multirow{4}{*}{{{COD10K}}}  
		& $ S_{\alpha}\uparrow $    & .626 & .697 & .804  & .815   & .810 & .809  & .817 & .811  & .822   \\
		& $ E_{\phi}\uparrow   $    & .601 & .691 & .870  & .880   & .875 & .876  & .881 & .882  & .888   \\
		& $\mathcal{M}\downarrow$   & .132 & .075 & .042  & .038   & .040 & .038  & .037 & .037  & .036   \\

	 \whline{1pt} 
	\end{tabular}\label{tab04}
    \end{center}
\end{table}

\textbf{Effectiveness of MFAM}. As shown in Table~\ref{tab04}, comparing ``B1" with ``A3", it can be observed that the model using MFAM can improve the COD performance. In the proposed MFAM, we adopt multi-scale convolutional kernels to extract the aggregated features from the original layers, which can capture multi-scale information to deal with the scale variations of camouflaged objects. Besides, comparing ``B2" with ``B1", it can be seen that our model using MFAM performs better than that using the Inception module in terms of most evaluation metrics.

\textbf{Effectiveness of CFPM}. Comparing ``C1" with ``B1", it can be seen that the feature fusion part in the CFPM can improve the COD performance, which indicates that the part effectively integrates the features from adjacent levels to exploit the correlations from cross-level features. Comparing ``C1" with ``C2", we can see that the feature propagation part further boosts the COD performance in CFPM. Therefore, combined with the two ablation experiments, the effectiveness of CFPM can be well validated. The proposed CFPM first carries out cross-level fusion and then conducts feature propagation, which can effectively propagate the useful information from the encoder to the decoder network for boosting the COD performance. Besides, comparing ``C2" with ``D", it can be observed that the proposed COD framework using CFPM performs better than that using the simple Attention Gate unit~\cite{oktay2018attention}. Therefore, the effectiveness of the proposed CFPM can be further validated. 

\textbf{Effectiveness of BGM}. Comparing ``C2" with ``E", it can be seen that the BGM can further improve the COD performance. This is mainly because the boundary-enhanced features are integrated into the decoder, which can provide much low-level structure information to locate the boundaries of camouflaged objects.

\begin{table}[tp]
  \centering
  \renewcommand{\arraystretch}{1.0}
  \renewcommand{\tabcolsep}{1.8mm}
  \caption{Results comparison for cross-dataset generalization. Our model is trained on one (rows) dataset and tested on all datasets (columns).} \vspace{-0.25cm}
  \scriptsize
  \begin{tabular}{c|c|cc|cc|cc}
  \whline{1pt} 
      \multirow{2}*{\textbf{Train Set}}   
    & \multirow{2}*{\textbf{Methods}}  
    & \multicolumn{2}{c|}{CHAMELEON}
    & \multicolumn{2}{c|}{CAMO-Test}
    & \multicolumn{2}{c}{COD10K-Test} \\

     \cline{3-8}
  
    & & $S_{\alpha}\uparrow$   &$\mathcal{M}\downarrow$
    & $S_{\alpha}\uparrow$   &$\mathcal{M}\downarrow$
    & $S_{\alpha}\uparrow$   &$\mathcal{M}\downarrow$ \\

    \whline{1pt}

    \multirow{5}{*}{{\rotatebox{90}{CAMO}}}    
     & MINet  & 0.803  & 0.080   & 0.777  & 0.113   & 0.713  & 0.089 \\
     & PraNet & 0.816  & 0.053   & 0.788  & 0.084   & 0.753  & 0.057 \\
     & SINet  & 0.803  & 0.066   & 0.756  & 0.103   & 0.728  & 0.068 \\
     & PFNet  & 0.809  & 0.061   & 0.784  & 0.087   & 0.736  & 0.064 \\
     & Ours   & 0.833  & 0.044   & 0.808  & 0.074   & 0.753  & 0.054 \\

    \whline{1pt}  

    \multirow{5}{*}{{\rotatebox{90}{COD10K}}}    
     & MINet  & 0.839  & 0.055   & 0.694  & 0.125   & 0.779  & 0.053 \\
     & PraNet & 0.867  & 0.035   & 0.688  & 0.117   & 0.802  & 0.038 \\
     & SINet  & 0.852  & 0.045   & 0.672  & 0.124   & 0.772  & 0.048 \\
     & PFNet  & 0.855  & 0.042   & 0.696  & 0.117   & 0.782  & 0.047 \\
     & Ours   & 0.874  & 0.034   & 0.708  & 0.111   & 0.820  & 0.036 \\

  \whline{1pt} 
  \end{tabular}\label{tab4}
\end{table}

\begin{table*}[tp]
  \centering
  
  \renewcommand{\arraystretch}{1.0}
  \renewcommand{\tabcolsep}{0.64mm}
  \caption{Quantitative polyp segmentation results on four widely datasets using five metrics (\ie, mDice, mIoU, $S_{\alpha}$~\cite{fan2017structure}, weighted $F_{\beta}$~\cite{achanta2009frequency}, and $\mathcal{M}$~\cite{perazzi2012saliency}). 
  ``$\uparrow$`` \& ``$\downarrow$'' indicate that larger or smaller is better. The best results are highlighted in \textbf{bold}. 
  } \vspace{-0.2cm}
  \scriptsize
  \begin{tabular}{r||ccccc|ccccc|ccccc|ccccc}
  \whline{1pt} 
    \multirow{2}*{\textbf{Methods}}   
    &\multicolumn{5}{c|}{CVC-ClinicDB \cite{bernal2015wm}}
    &\multicolumn{5}{c|}{ETIS~\cite{silva2014toward}}
    &\multicolumn{5}{c|}{CVC-ColonDB~\cite{tajbakhsh2015automated}}
    &\multicolumn{5}{c}{Kvasir~\cite{jha2020kvasir}}\\

     \cline{2-21}
  
    & mDice $\uparrow$  & mIou $\uparrow$  & $S_{\alpha}\uparrow$  &$F_{\beta}^{w}\uparrow$  &$\mathcal{M}\downarrow$
    & mDice $\uparrow$  & mIou $\uparrow$  & $S_{\alpha}\uparrow$  &$F_{\beta}^{w}\uparrow$  &$\mathcal{M}\downarrow$
    & mDice $\uparrow$  & mIou $\uparrow$  & $S_{\alpha}\uparrow$  &$F_{\beta}^{w}\uparrow$  &$\mathcal{M}\downarrow$
    & mDice $\uparrow$  & mIou $\uparrow$  & $S_{\alpha}\uparrow$  &$F_{\beta}^{w}\uparrow$  &$\mathcal{M}\downarrow$ \\

    \whline{1pt}  
  
  
    U-Net~\cite{ronneberger2015u}~
    & 0.823   & 0.755   & 0.889   & 0.811   & 0.019
    & 0.398   & 0.335   & 0.684   & 0.366   & 0.036	
    & 0.512   & 0.444   & 0.712   & 0.498   & 0.061
    & 0.818   & 0.746   & 0.858   & 0.794   & 0.056\\
    
    U-Net++~\cite{zhou2018unet++}~
    & 0.794   & 0.729   & 0.873   & 0.785   & 0.022
    & 0.401   & 0.344   & 0.683   & 0.390   & 0.035
    & 0.483   & 0.410   & 0.691   & 0.467   & 0.064
    & 0.821   & 0.743   & 0.862   & 0.808   & 0.048	\\

    SFA~\cite{fang2019selective}~
    & 0.723   & 0.611   & 0.782   & 0.670   & 0.075
    & 0.297   & 0.217   & 0.557   & 0.231   & 0.109	
    & 0.469   & 0.347   & 0.634   & 0.379   & 0.094
    & 0.723   & 0.611   & 0.782   & 0.670   & 0.075\\    

    PraNet~\cite{fan2020pranet}~
    & 0.899 & 0.849 & 0.936 & 0.896   & 0.009
    & 0.628 & 0.567 & 0.794 & 0.600   & 0.031 
    & 0.709 & 0.640 & 0.819 & 0.696   & 0.045
    & 0.898 & 0.840 & 0.915 & 0.885   & 0.030\\   
    \hline
    Ours~
    & \textbf{0.925}   & \textbf{0.877}   & \textbf{0.947}   & \textbf{0.910}   & \textbf{0.008}
    & \textbf{0.717}   & \textbf{0.643}   & \textbf{0.841}   & \textbf{0.657}   & \textbf{0.019}  
    & \textbf{0.731}   & \textbf{0.658}   & \textbf{0.831}   & \textbf{0.735}   & \textbf{0.038}
    & \textbf{0.902}   & \textbf{0.849}   & \textbf{0.919}   & \textbf{0.894}   & \textbf{0.027}\\

  \whline{1pt} 
  \end{tabular}\label{tab5}
\end{table*}

\subsection{Cross-dataset Generalization}

The cross-dataset generalization study plays a crucial role in assessing different algorithms. Here, we utilize the cross-dataset analysis strategy \cite{torralba2011unbiased} to evaluate the generalizability of our model, \ie, training a model on one dataset and then testing it on others. To investigate the generalization ability of our model and other SOTA methods, we train the proposed methods and four SOTA methods (\ie, MINet~\cite{pang2020multi}, PraNet~\cite{fan2020pranet}, SINet~\cite{fan2020camouflaged}, and PFNet~\cite{mei2021camouflaged}) on the CAMO and COD1K datasets, and then report the results on the test sets. Table~\ref{tab4} shows the comparison of results for cross-dataset generalization. As shown in Table~\ref{tab4}, it can be seen that our method still performs better than other comparison methods under two different training sets. Besides, comparing the results between Table~\ref{tab1} and Table~\ref{tab4}, it can be observed that all methods drop the performance when training one dataset and testing on the other datasets.

\subsection{Extension Application}

Automatic polyp segmentation is an important step in modern polyp screening systems, which can help clinicians accurately locate polyp regions for further diagnosis or treatments. Similar to camouflaged object detection, polyp segmentation also faces several challenges, including 1) variations in the shape and size of polyps and 2) non-sharp boundary between a polyp and its surrounding mucosa \cite{fan2020pranet}. Therefore, to further validate the effectiveness of our \ours, we extend it to the polyp segmentation task.

\textbf{Experimental Settings}. The comparison experiments are conducted on four polyp segmentation datasets, which are CVC-ClinicDB \cite{bernal2015wm},  ETIS~\cite{silva2014toward},  CVC-ColonDB~\cite{tajbakhsh2015automated}, and  Kvasir~\cite{jha2020kvasir}. To validate the effectiveness of our model on the polyp segmentation task, we compare the proposed model with four state-of-the-art methods, \ie, UNet \cite{ronneberger2015u}, UNet++~\cite{zhou2018unet++}, SFA~\cite{fang2019selective}, and PraNet~\cite{fan2020pranet}. Besides, following the same setting in~\cite{fan2020pranet}, we train the proposed model on the CVC-ClinicDB and Kvasir datasets. Moreover, we utilize five widely used metrics, \ie, mean Dice coefficient (mDice), mean Intersection over Union (mIoU), $S_{\alpha}$~\cite{fan2017structure}, $F_{\beta}^{w}$~\cite{achanta2009frequency}, and $\mathcal{M}$~\cite{perazzi2012saliency}, for quantitative evaluation.

\textbf{Results Comparison}. Table~\ref{tab5} shows the quantitative and qualitative results on four polyp datasets. From the results in Table~\ref{tab5}, it can be seen that our model performs best than the four compared methods and improve the polyp segmentation performance by a large margin. Specifically, on the CVC-ColonDB dataset, our model achieves $2.9\%$, $3.3\%$, $1.2\%$, and $1.6\%$ improvements over PraNet in the terms of mDice, mIou, $S_{\alpha}$, and $F_{\beta}^{w}$, respectively. On the ETIS dataset, our model achieves $14.2\%$, $13.4\%$, $5.9\%$, and $9.5\%$ improvements over PraNet in the terms of mDice, mIou, $S_{\alpha}$, and $F_{\beta}^{w}$, respectively. Therefore, the results validate the effectiveness of the proposed model can effectively segment polyps in a complex background. Besides, Fig.~\ref{fig07} shows a visual comparison of the four polyp segmentation methods. From the results, it can be observed that the proposed method can accurately locate and segment the polyps in several challenging factors, such as varied size, non-sharp boundary, etc.

\section{Conclusion}
\label{conclusion}

In this paper, we have proposed a novel camouflaged object detection framework, \ie, Feature Aggregation and Propagation Network (\ours). We first utilize the proposed BGM to explicitly model the boundary characteristic, and the obtained boundary-enhanced feature representations are integrated into the decoder to boost the COD performance. Then, we propose the MFAM to extract the multi-scale information from a single layer for dealing with scale variations. Moreover, we design the CFPM to effectively fuse cross-level features and then propagate them to the decoder network with the valuable context information from the encoder. Extensive experimental results demonstrate that our \ours~outperforms other state-of-the-art COD methods. Furthermore, we apply \ours~to the polyp segmentation task, and the results show that our model outperforms other polyp segmentation methods.

\ifCLASSOPTIONcaptionsoff
  \newpage
\fi

\bibliographystyle{IEEEtran}
\bibliography{ref}

\end{document}